\documentclass[review]{elsarticle}

\usepackage{graphicx}
\usepackage{epstopdf, epsfig}
\usepackage{subfigure}
\usepackage{placeins}
\usepackage{mathrsfs}

\usepackage{amsmath}
\usepackage{amssymb}
\usepackage{upgreek}

\usepackage{fullpage}
\usepackage{hyperref}

\biboptions{authoryear}

\begin{document}
	
	\begin{frontmatter}
		
		\title{Distributed physics informed neural network for data-efficient solution to partial differential equations}
		
		\author[mymainaddress]{Vikas Dwivedi\corref{mycorrespondingauthor}}
		\cortext[mycorrespondingauthor]{Corresponding author}
		\ead{me15d080@smail.iitm.ac.in}
		
		\author[mysecondaryaddress]{Nishant Parashar}
		\ead{nishantparashar14@gmail.com}
		
		\author[mymainaddress]{Balaji Srinivasan}
		\ead{balaji.srinivasan@gmail.com}

		\address[mymainaddress]{Department of Mechanical Engineering, Indian Institute of Technology Madras,
			Chennai 600036, India}
		
		\address[mysecondaryaddress]{Department of Applied Mechanics, Indian Institute of Technology Delhi, New Delhi 110016, India}

		
		\begin{abstract}
			The physics informed neural network (PINN) is evolving as a viable method to solve partial differential equations. In the recent past PINNs have been successfully tested and validated to find solutions to both linear and non-linear partial differential equations (PDEs). However, the literature lacks detailed investigation of PINNs in terms of their representation capability. In this work, we first test the original PINN method in terms of its capability to represent a complicated function. Further, to address the shortcomings of the PINN architecture, we propose a novel distributed PINN, named DPINN. We first perform a direct comparison of the proposed DPINN approach against PINN to solve a non-linear PDE (Burgers' equation). We show that DPINN not only yields a more accurate solution to the Burgers' equation, but it is found to be more data-efficient as well. At last, we employ our novel DPINN to two-dimensional steady-state Navier-Stokes equation, which is a system of non-linear PDEs. To the best of the authors' knowledge, this is the first such attempt to directly solve the Navier-Stokes equation using a physics informed neural network. 
		\end{abstract}
		
		
	\end{frontmatter}
	
	
	\section{Introduction}\label{s:intro}
    Partial differential equations (PDEs) are extensively used in the mathematical modelling of a variety of problems in physics, engineering and finance. For a variety of practical problems of interest, the analytical solution of these PDEs is generally unknown. Over the years, various numerical methods have been developed to solve these PDEs.  Most of the popular numerical methods viz. finite element method \cite[]{rao2017finite}, finite difference method \cite[]{leveque2007finite} and finite volume method \cite[]{versteeg2007introduction} are mesh-based methods. Although these numerical methods are very successful and are widely used, there are various problems associated with these methods. For example, it is a well-known fact that mesh generation is a very difficult task with complex geometries. Furthermore, the discretisation of the PDE itself introduces truncation errors, which can be a quite serious problem \cite[]{quirk1997contribution}. 
    
    With the recent advances in the computational resources and availability of data,  the deep neural networks have evolved as a viable method to solve PDEs. The performance of a neural network in solving PDEs depends on two main factors: (1) neural network architecture and (2) learning algorithms. As a result, the design of new neural network architectures and learning algorithms has become an active area of research for the past decade. Recently, \cite[]{berg2018unified} have produced good results in solving stationary PDEs with complex geometries and  \cite{sirignano2018dgm} have introduced a deep Galerkin method to solve high dimensional PDEs. We also refer to the earlier works of \cite{lagaris1998, lagaris2000} in which they solved the initial boundary value problem using neural networks and later extended their work to handle irregular boundaries. Since then, a lot of researchers have made their contributions in solving initial and boundary value problems in arbitrary boundaries \cite[]{mcfall2009artificial,kumar2011multilayer,mall2016application}.  
    
    Among all the deep neural network-based approaches to solve PDEs, we particularly refer to the physics-informed learning machines which have shown promising results for a series of nonlinear benchmark problems. The peculiar property of this approach is the inclusion of the prior-structured information about the solution in the learning algorithm. The initial promising results using this method were reported by \cite[]{owhadi2015bayesian,raissi2017inferring,raissi2017machine}. The authors employed the Gaussian process regression to accurately infer the solutions to linear problems along with the associated uncertainty estimates. Further, their method was extended to nonlinear problems by \cite{raissi2018hidden} and \cite{raissi2018numerical}  in the context of both inference and system identification. Finally, the physics informed neural networks (PINN) are introduced by \cite{raissi2019physics} for both data-driven solution of PDEs as well as the data-driven discovery of parameters in a PDE.
    
    Although PINNs have been successfully tested to retrieve accurate solutions for a wide variety of PDEs, they suffer from several drawbacks \cite[]{raissi2019physics}. For example (i) the required depth of the PINN increases with increasing order of PDE leading to slow learning-rate due to the well-known issue of vanishing gradients, (ii) PINNs are not robust in representing sharp local gradients in a broad computational domain \cite[]{2019arXiv190703507D}. In addition to this, there is a lot of uncertainty in terms of requirement of the amount of training data and the number of hidden layers for efficient implementation of the PINN.    
    
    In this paper, we propose a distributed version of the original PINN called distributed PINN (or DPINN). The DPINN handles several issues encountered by the original PINN \cite[]{raissi2019physics} by efficiently designing the network architecture and modifying the associated cost function. In the DPINN approach, the computational domain is distributed into smaller sub-domains (called cells), and simpler PINNs are employed in each of these individual cells. These individual PINNs (in the comprehensive DPINN) are just two layers deep. Hence, solving the potential vanishing gradients issue of using deeper PINNs.
    Although this partitioning complicates the cost function as we have to penalise for the interfacial flux mismatch across the neighbouring cells, still it simplifies the function or PDE to be represented by the local PINN. This makes DPINN more data-efficient in comparison to the original PINN. 
    
    This paper is organised into five sections. In Section \ref{s:pinn} we, present a brief overview of the physics informed neural network (PINN) of \cite{raissi2019physics}. The proposed novel DPINN architecture, along with the mathematical formulation, is described in Section \ref{s:dpinn}. The detailed evaluation of DPINN, in terms of its representation capability and the ability to solve non-linear PDEs in performed in Section \ref{s:results}. Section \ref{s:summary} concludes the work with a summary.
	
	\section{Review of physics informed neural network}
	\label{s:pinn}
	\cite{raissi2019physics} proposed a data-efficient PINN network for approximating solutions to general non-linear PDEs and validated it with a series of benchmark test cases. The main feature of the PINN is the inclusion of the prior knowledge of physics in the learning algorithm as cost function. As a result, the algorithm imposes a penalty for any non-physical solution
	and quickly directs it towards the correct solution. This physics
	informed approach enhances the information content of the data. As a result, the algorithm has good generalization property even in the small data set regime. 
	
	\subsubsection{Mathematical formulation}
	\label{ss:pinn_math}
	Consider a PDE of the following form:
	\begin{equation}
	\frac{\partial}{\partial t}u(\overrightarrow{x},t)+\mathcal{\mathscr{N}}u (\overrightarrow{x},t)=R(\overrightarrow{x},t),\;(\overrightarrow{x},t)\epsilon\varOmega\text{x}[0,T],\label{eq:gen_PDE}
	\end{equation}
	\begin{equation}
	u(\overrightarrow{x},t)=B(\overrightarrow{x},t),\;(\overrightarrow{x},t)\epsilon\partial\varOmega\text{x}[0,T],\label{eq:gen_BC}
	\end{equation}
	\begin{equation}
	u(\overrightarrow{x},0)=F(\overrightarrow{x}),\;\overrightarrow{x}\epsilon\varOmega,\label{eq:gen_IC}
	\end{equation}
	where $\mathcal{\mathscr{N}}$ may be a linear or nonlinear differential
	operator and $\partial\varOmega$ is the boundary of computational
	domain $\Omega$. We approximate $u(\overrightarrow{x},t)$ with the
	output \textbf{$f(\overrightarrow{x},t)$} of the neural network.
	\begin{figure}
		\center
		\includegraphics[scale=0.5]{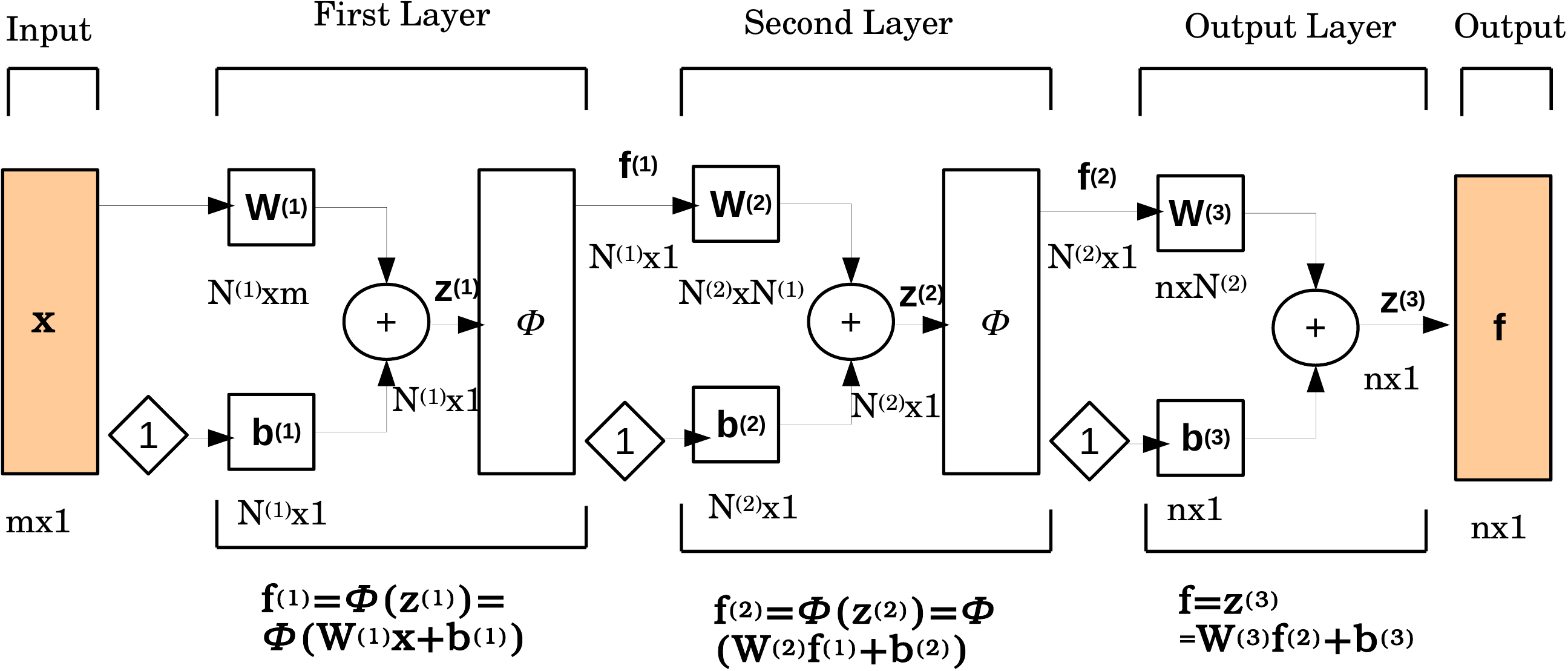}
		\caption{\label{fig:PINN_SCHEMATIC}Schematic of a minimal deep neural network}
	\end{figure}
	For simplicity, a schematic diagram of a simple two layer deep neural network with an $m\text{x}1$ input $\overrightarrow{x}$ and an $n\text{x}1$ output $\overrightarrow{f}$ is shown in Fig \ref{fig:PINN_SCHEMATIC}.
	In this case, assuming $n=1$, the neural network output is given by:
	\begin{equation}
	f=\boldsymbol{W}^{(3)}\phi(\boldsymbol{W}^{(2)}(\phi(\boldsymbol{W}^{(1)}\overrightarrow{x}+\overrightarrow{b}^{(1)}))+\overrightarrow{b}^{(2)})+\overrightarrow{b}^{(3)}
	\end{equation}
	where $\phi$ is the activation function and $\boldsymbol{W}^{(k)}$
	and $\overrightarrow{b}^{(k)}$ represent the weight matrix and bias vector of $k^{th}$ layer. The expression for deeper networks can also be written in a similar fashion. 
	The essence of PINN lies in the definition of its loss function. In
	order to make the neural network ``physics informed'', the loss function is defined such that a penalty is imposed whenever the network output doesn't respect the physics of the problem. If we denote the
	training errors in approximating the PDE, BCs and IC by $\overrightarrow{\xi}_{f}$,
	$\overrightarrow{\xi}_{bc}$ and $\overrightarrow{\xi}_{ic}$ respectively.
	Then, the expressions for these errors are as follows:
	
	\begin{equation}
	\overrightarrow{\xi}_{f}=\frac{\partial\overrightarrow{f}}{\partial t}+\mathcal{\mathcal{\mathscr{N}}}\overrightarrow{f}-\overrightarrow{R},\;(\overrightarrow{x},t)\epsilon\varOmega\text{x}[0,T],\label{eq:PDE_ERROR}
	\end{equation}
	\begin{equation}
	\overrightarrow{\xi}_{bc}=\overrightarrow{f}-\overrightarrow{B,}\;(\overrightarrow{x},t)\epsilon\partial\varOmega\text{x}[0,T],
	\end{equation}
	
	\begin{equation}
	\overrightarrow{\xi}_{ic}=\overrightarrow{f}(.,0)-\overrightarrow{F},\;\overrightarrow{x}\epsilon\varOmega.\label{eq:IC_ERROR}
	\end{equation}
	The expressions for $\frac{\partial\overrightarrow{f}}{\partial t}$
	and $\mathcal{\mathcal{\mathscr{N}}}\overrightarrow{f}$ used in the
	equation (\ref{eq:PDE_ERROR}) can be computed using automatic differentiation \cite[]{baydin2018automatic}.
	The loss function $J$ to be minimized for a PINN is given by
	\begin{equation}
	J=J_{PDE}+J_{BC}+J_{IC}
	\end{equation}
	where $J_{PDE}$, $J_{BC}$ and $J_{IC}$ correspond to losses at
	collocation, boundary condition and initial condition data respectively.
	The expressions for these losses are given below:
	\begin{equation}
	J_{PDE}=\frac{\overrightarrow{\xi}_{f}^{T}\overrightarrow{\xi}_{f}}{2N_{f}},
	\end{equation}
	\begin{equation}
	J_{BC}=\frac{\overrightarrow{\xi}_{bc}^{T}\overrightarrow{\xi}_{bc}}{2N_{bc}},
	\end{equation}
	\begin{equation}
	J_{IC}=\frac{\overrightarrow{\xi}_{ic}^{T}\overrightarrow{\xi}_{ic}}{2N_{ic}},
	\end{equation}
	where $N_{f}$, $N_{bc}$ and $N_{ic}$ refer to number of collocation
	points, boundary condition points and initial condition points respectively. Finally, any gradient-based optimization routine may be used to minimize $J$. This completes the mathematical formulation of PINN. The key steps in its implementation are as follows:
	\begin{enumerate}
		\item Identify the PDE to be solved along with the initial and boundary
		conditions. 
		\item Decide the architecture of PINN.
		\item Approximate the correct solution with PINN.
		\item Find expressions for the PDE, BCs and IC in terms of PINN and its
		derivatives. 
		\item Define a loss function which penalizes for error in PDE, BCs and IC.
		\item Minimize the loss with gradient-based algorithms.
	\end{enumerate}
	
	\section{Distributed PINN}
	\label{s:dpinn}
	In this section, we propose a distributed version of PINN called DPINN. This algorithm takes motivation from finite volume methods in which the whole computational domain is partitioned into multiple cells, and the governing equations are solved for each cell. The solutions of these individual cells are stitched together with additional convective
	and diffusive fluxes conditions at the cell interfaces. We adopt a similar strategy in DPINN. As the representation of a complex function is very hard for a single PINN in the whole domain, we divide the domain into multiple cells and install a PINN in each cell. Therefore, each PINN uses different representations in different portions of the domain while satisfying some additional constraints of continuity and differentiability. In the conventional PINNs, the prior information of the physics of the problem acts as a regularizer in a global sense.
	However in the DPINN approach, the additional interface conditions act as local regularization agents that further constrain the space of admissible solution.
	
	\subsection{Mathematical formulation}
	\label{ss:dpinn_math}
	\begin{figure}
		\center
		\subfigure[Distributed PINNs in $\Omega$]{\begin{raggedright}
				\includegraphics[scale=0.3]{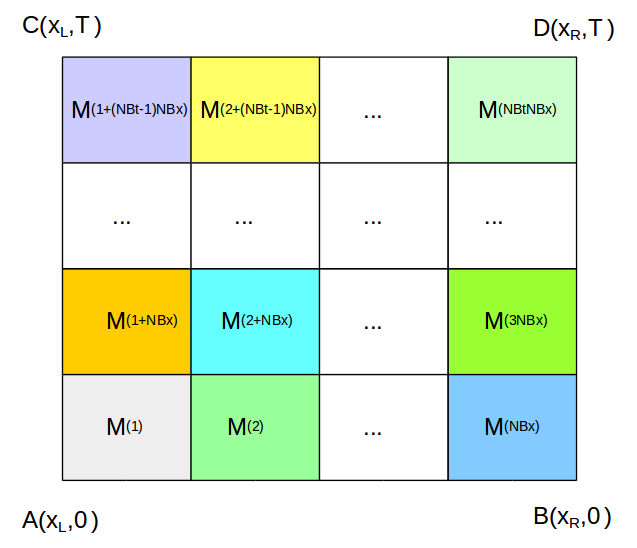}
				\par\end{raggedright}}
		\subfigure[A PINN in $\Omega_{i}$. Red triangles: boundary points, green rectangles:
		collocation points ]{\begin{raggedleft}
				\includegraphics[scale=0.3]{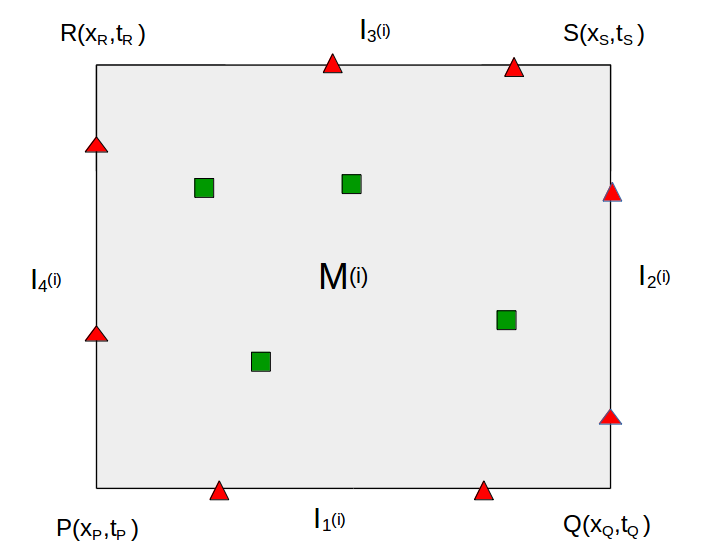}
				\par\end{raggedleft}}
		\caption{\label{fig:DPIELM in 2D}DPINN architecture for the full domain and
			an individual cell. }
	\end{figure}
	Consider the following 1D unsteady problem
	\begin{equation}
	\frac{\partial}{\partial t}u(x,t)+\mathcal{\mathcal{\mathscr{N}}}u(x,t)=R(x,t),(x,t)\epsilon\varOmega
	\end{equation}
	\begin{equation}
	u(x,t)=B(x,t),(x,t)\epsilon\partial\varOmega,
	\end{equation}
	
	\begin{equation}
	u(x,0)=F(x),x\epsilon[x_{L},x_{R}],
	\end{equation}
	where $\mathcal{\mathscr{N}}$ is a potentially nonlinear differential
	operator and $\partial\varOmega$ is the boundary of computational
	domain $\Omega$. In this problem, the rectangular domain $\Omega$
	is given by $\Omega=[x_{L},x_{R}]\text{x}[0,T]$. On uniformly dividing
	$\Omega$ into $N_{c}$ non-overlapping rectangular cells, $\Omega$
	may be written as
	\begin{equation}
	\Omega=\bigcup_{i=1}^{N_{c}}\Omega_{i}.
	\end{equation}
	The boundary of the cell $\Omega_{i}$ is denoted by $\partial\varOmega_{i}$.
	For rectangular cells,
	\begin{equation}
	\partial\Omega_{i}=\bigcup_{m=1}^{4}I_{m}^{(i)}
	\end{equation}
	where $I_{m}^{(i)}$ represents the $m^{th}$ interface of $\Omega_{i}$. 
	
	Fig (\ref{fig:DPIELM in 2D}a) shows the distribution of PINNs in
	a rectangular computational domain with $NB_{x}\text{x}NB_{t}$ cells
	(i.e. $N_{c}=NB_{x}\text{x}NB_{t}$). We denote the PINN on the $i^{th}$
	cell by $M^{(i)}$. Fig (\ref{fig:DPIELM in 2D}b) shows a PINN with
	collocation points at the interior and the boundary points at the
	four interfaces. Each individual PINN has its own set of weights and
	biases. The output corresponding to a given $M^{(i)}$ is denoted
	by $f^{(i)}$. 
	
\begin{figure}
		\center
		\includegraphics[scale=0.45]{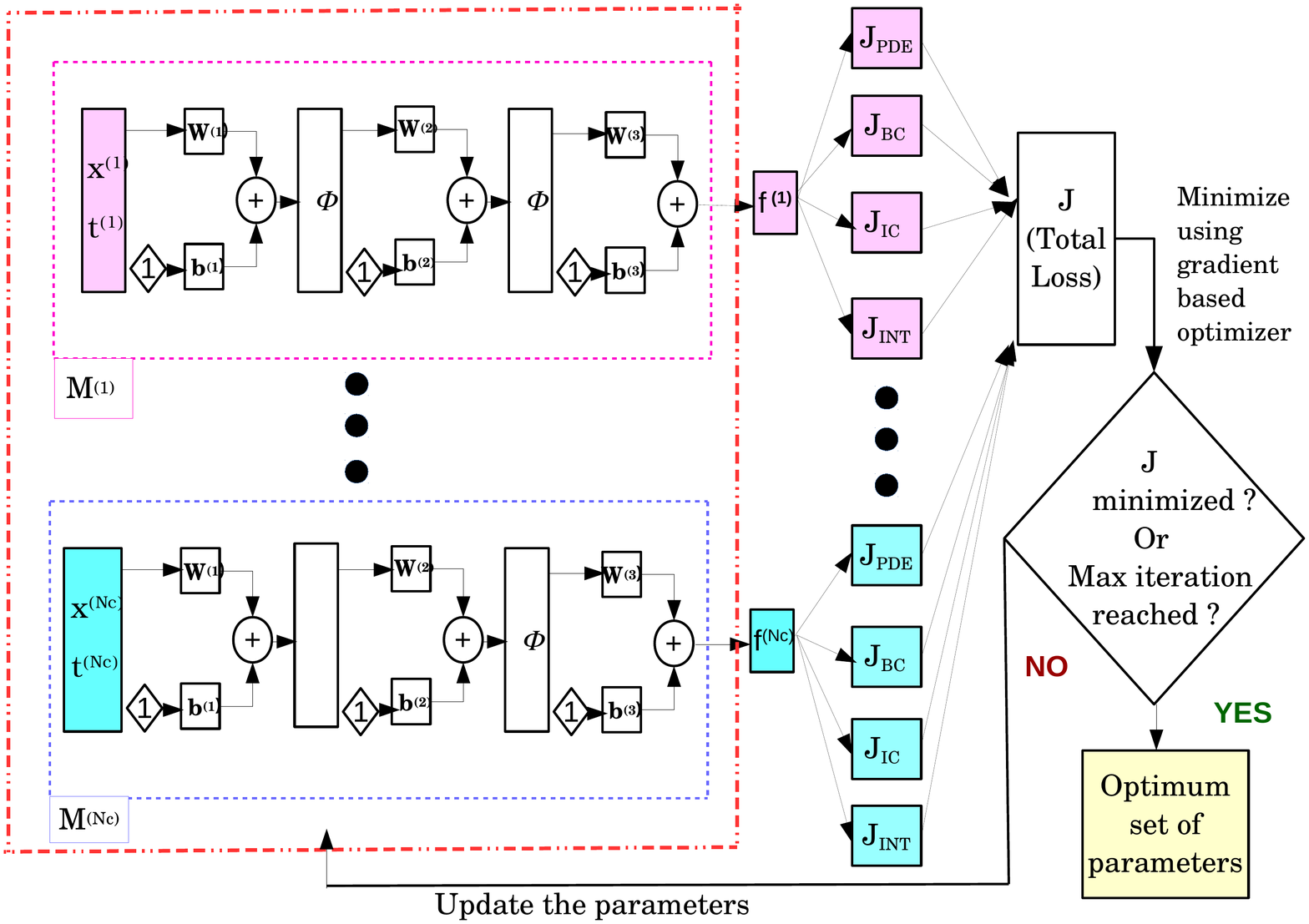}
		\caption{\label{fig:DPINN_SCHEMATIC}Schematic of DPINN architecture}
	\end{figure}

	At each $M^{(i)}$, we enforce additional constraints of continuity (or smoothness) of solution at the cell interfaces depending on the differential operator $\mathcal{\mathscr{N}}$. For example, continuity
	of solution is sufficient for advection problems. For the diffusion problem, the solution should be continuously differentiable. 
	For the computational domain shown in figure (\ref{fig:DPIELM in 2D}), the loss function to be minimized is a combination of conventional PINN losses and additional interface losses and is given by:
	\begin{equation}
	J=J_{PDE}+J_{BC}+J_{IC}+J_{interface}.
	\end{equation}
	Figure \ref{fig:DPINN_SCHEMATIC} presents a detailed schematic of the DPINN architecture.

	\subsubsection{Expressions for the conventional PINN losses  }
	
	\begin{equation}
	J_{PDE}=\sum\frac{\overrightarrow{\xi}_{f}^{(i)^{T}}\overrightarrow{\xi}_{f}^{(i)}}{2N_{f}^{(i)}},\label{eq:DPIELM_start}
	\end{equation}
	
	\begin{equation}
	J_{BC}=\sum\frac{\overrightarrow{\xi}_{bc}^{(j)}\overrightarrow{\xi}_{bc}^{(j)}}{2N_{bc}^{(j)}}+\sum\frac{\overrightarrow{\xi}_{bc}^{(k)^{T}}\overrightarrow{\xi}_{bc}^{(k)}}{2N_{bc}^{(k)}},
	\end{equation}
	\begin{equation}
	J_{IC}=\sum\frac{\overrightarrow{\xi}_{ic}^{(l)}\overrightarrow{\xi}_{ic}^{(l)}}{2N_{ic}^{(l)}},
	\end{equation}
	where $\overrightarrow{\xi}_{f}^{(i)}=\left(\frac{\partial\overrightarrow{f}^{(i)}}{\partial t}+\mathcal{L}\overrightarrow{f}^{(i)}-\overrightarrow{R}^{(i)}\right)_{\Omega_{i}}$,
	$\overrightarrow{\xi}_{bc}^{(j)}=(\overrightarrow{f}^{(j)}-\overrightarrow{B}^{(j)})_{I_{4}}$,
	$\overrightarrow{\xi}_{bc}^{(k)}=(\overrightarrow{f}^{(k)}-\overrightarrow{B}^{(k)})_{I_{2}}$,
	$\overrightarrow{\xi}_{ic}^{(l)}=(\overrightarrow{f}^{(l)}-\overrightarrow{B}^{(l)})_{I_{1}}$,
	$i=[1,2,...,N_{c}]$, $j=[1,(1+NB_{x}),...,(1+(NB_{t}-1)NB_{x})]$,
	$k=[NB_{x},2NB_{x},...,NB_{t}\text{x}NB_{x}]$ and $l=[1,2,...,NB_{x}]$.
	
	\subsubsection{Expressions for the additional interface losses }
	
	In general, we consider total interface loss as sum of continuity
	and differentiability losses i.e. 
	\begin{equation}
	J_{interface}=J_{C_{x}^{0}}+J_{C_{t}^{0}}+J_{C_{x}^{1}}
	\end{equation}
	where $J_{C_{x}^{0}}$, $J_{C_{t}^{0}}$ refer to continuity losses
	along $x$ and $t$ interfaces and $J_{C_{x}^{1}}$ refers to differentiability loss along $x$ interfaces. The expressions for these losses are as follows:
	\begin{equation}
	J_{C_{x}^{0}}=\sum\frac{\overrightarrow{\xi}_{C_{x}^{0}}^{(i)^{T}}\overrightarrow{\xi}_{C_{x}^{0}}^{(i)}}{2N_{C^{0}}^{(i)}},
	\end{equation}
	\begin{equation}
	J_{C_{t}^{0}}=\sum\frac{\overrightarrow{\xi}_{C_{t}^{0}}^{(i)^{T}}\overrightarrow{\xi}_{C_{t}^{0}}^{(i)}}{2N_{C^{0}}^{(i)}},
	\end{equation}
	\begin{equation}
	J_{C_{x}^{1}}=\sum\frac{\overrightarrow{\xi}_{C_{x}^{1}}^{(i)^{T}}\overrightarrow{\xi}_{C_{x}^{1}}^{(i)}}{2N_{C_{x}^{1}}^{(i)}},
	\end{equation}
	where 
	\[
	\overrightarrow{\xi}_{C_{x}^{0}}^{(i)}=\left\{ \begin{array}{c}
	\overrightarrow{f}^{(\kappa_{x}^{0}(1)+i-1)}\\
	\overrightarrow{f}^{(\kappa_{x}^{0}(2)+i-1)}\\
	...\\
	\overrightarrow{f}^{(\kappa_{x}^{0}(NB_{t})+i-1)}
	\end{array}\right\} _{I_{2}}-\left\{ \begin{array}{c}
	\overrightarrow{f}^{(\kappa_{x}^{0}(1)+i)}\\
	\overrightarrow{f}^{(\kappa_{x}^{0}(2)+i)}\\
	...\\
	\overrightarrow{f}^{(\kappa_{x}^{0}(NB_{t})+i)}
	\end{array}\right\} _{I_{4}},
	\]
	\[
	\overrightarrow{\xi}_{C_{t}^{0}}^{(j)}=\left\{ \begin{array}{c}
	\overrightarrow{f}^{(\kappa_{t}^{0}(1)+(j-1)NB_{x})}\\
	\overrightarrow{f}^{(\kappa_{t}^{0}(2)+(j-1)NB_{x})}\\
	...\\
	\overrightarrow{f}^{(\kappa_{t}^{0}(NB_{x})+(j-1)NB_{x})}
	\end{array}\right\} _{I_{3}}-\left\{ \begin{array}{c}
	\overrightarrow{f}^{(\kappa_{t}^{0}(1)+jNB_{x})}\\
	\overrightarrow{f}^{(\kappa_{t}^{0}(2)+jNB_{x})}\\
	...\\
	\overrightarrow{f}^{(\kappa_{t}^{0}(NB_{x})+jNB_{x})}
	\end{array}\right\} _{I_{1}},
	\]
	\[
	\overrightarrow{\xi}_{C_{x}^{1}}^{(k)}=\frac{\partial}{\partial x}\left\{ \begin{array}{c}
	\overrightarrow{f}^{(\kappa_{x}^{0}(1)+i-1)}\\
	\overrightarrow{f}^{(\kappa_{x}^{0}(2)+i-1)}\\
	...\\
	\overrightarrow{f}^{(\kappa_{x}^{0}(NB_{t})+i-1)}
	\end{array}\right\} _{I_{2}}-\frac{\partial}{\partial x}\left\{ \begin{array}{c}
	\overrightarrow{f}^{(\kappa_{x}^{0}(1)+i)}\\
	\overrightarrow{f}^{(\kappa_{x}^{0}(2)+i)}\\
	...\\
	\overrightarrow{f}^{(\kappa_{x}^{0}(NB_{t})+i)}
	\end{array}\right\} _{I_{4}},
	\]
	
	$\kappa_{x}^{0}=[1,(1+NB_{x}),...,1+(NB_{t}-1)NB_{x}]^{T}$, $\kappa_{t}^{0}=[1,2,...,NB_{x}]^{T},$
	$i=[1,2,...,NB_{x}-1]$ and $j=[1,2,...,NB_{t}-1].$
	
	It is to be noted that although we have shown the formulation for 1D unsteady problems, no special adjustment is needed to extend the formulation to higher dimensional problems.
	This completes the mathematical formulation of DPINN. The main steps in its implementation are as follows:
	\begin{enumerate}
		\item Divide the computational domain into uniformly distributed non-overlapping
		cells and install a PINN in each cell.
		\item Depending on the cell location, PDE and the initial and boundary conditions,
		find the conventional PINN losses at each cell.
		\item Depending on the PDE, calculate interface losses at each cell interface.
		\item Calculate the total loss by taking the summation of all the losses. 
		\item Minimize the total loss using a gradient descent algorithm.
	\end{enumerate}

	
	\section{Results and Discussions}
	\label{s:results}
	In this section, we first test the original PINN developed by \cite{raissi2019physics} in terms of its ability to represent complicated functions. In this regard, we employ the PINN to solve the advection equation with complex initial conditions. After that, we discuss the bottlenecks of the PINN and explain how the proposed DPINN is an improvement over the PINN. Further, we evaluate the performance of DPINNs in solving non-linear PDEs as well. We first test our network for predicting the solution to the Burgers' equation. The performance is evaluated in terms of accuracy as compared to numerically accurate solutions and the PINN of \cite{raissi2019physics}. After that, we evaluate our model on a system of non-linear PDEs, namely, the Navier Stokes equations and validate the results against well-established results of \cite{ghia1982high}. 
	
	For all the problems presented in this work, we have used the $tanh$ non-linear activation function. The Adam optimizer \cite[]{Adam},  is kept as the default optimizer throughout this work. The learning rate is 0.001 for advection and Burgers' equation and 0.0001 for the Navier Stokes equation. The input data in all of the individual cells of the DPINN is normalised in the range $\left[0,1\right]$. 
	
	\subsection{Advection equation}
	\label{ss:advection}
\begin{figure*}[h]
		\begin{center}
			\subfigure[]{
				\resizebox*{5.31cm}{!}{\includegraphics{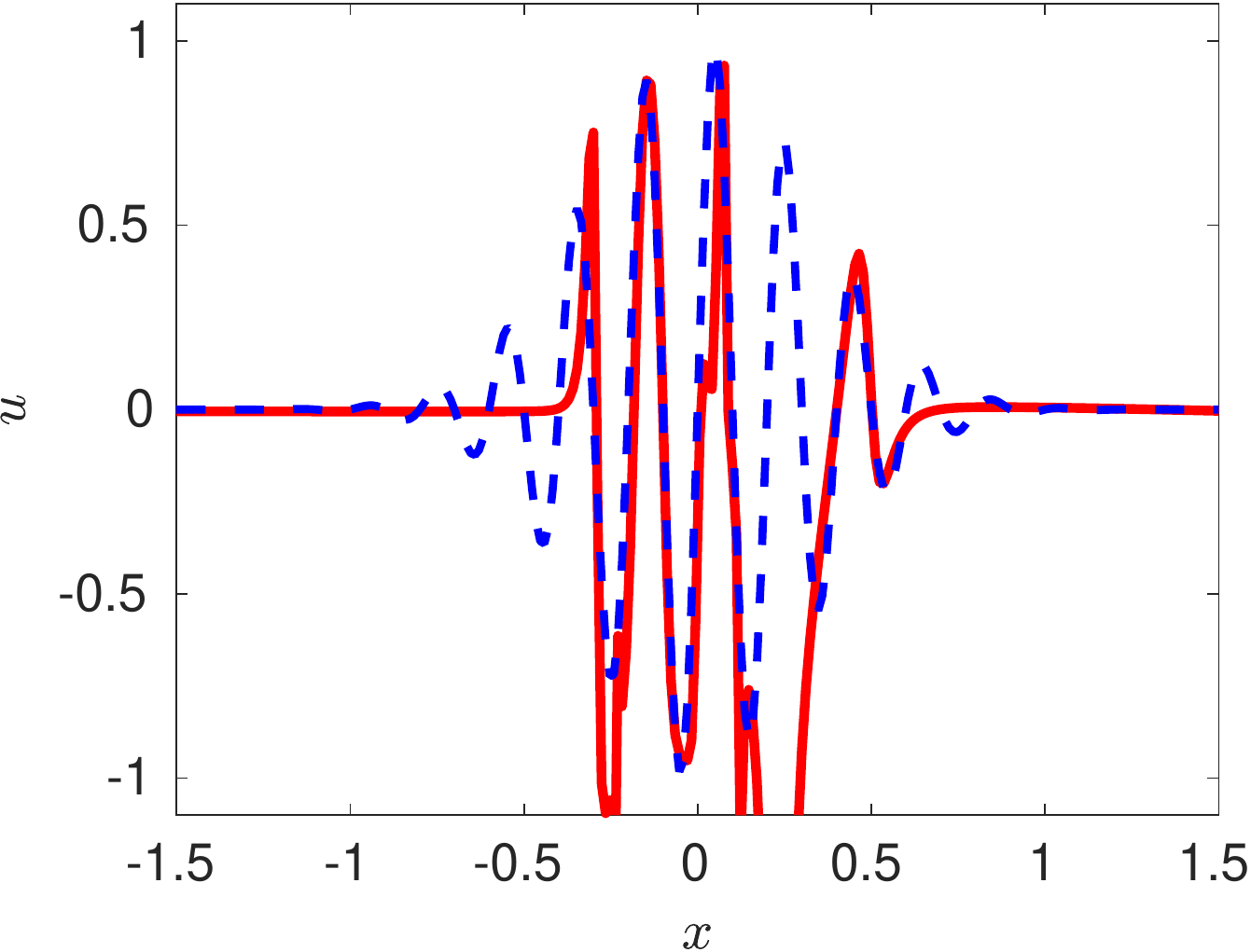}}}
			\subfigure[]{
				\resizebox*{5.31cm}{!}{\includegraphics{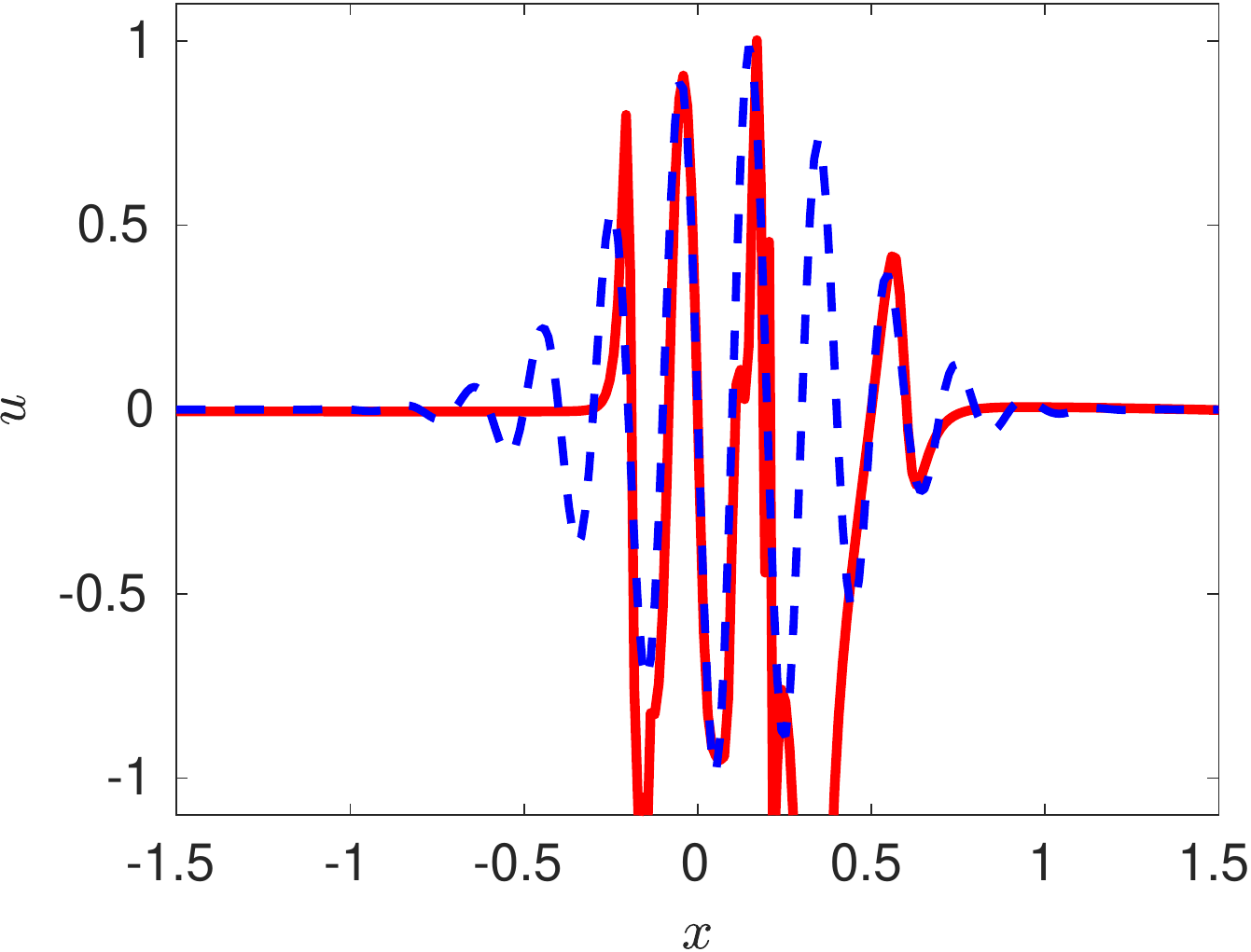}}}
			\subfigure[]{
				\resizebox*{5.31cm}{!}{\includegraphics{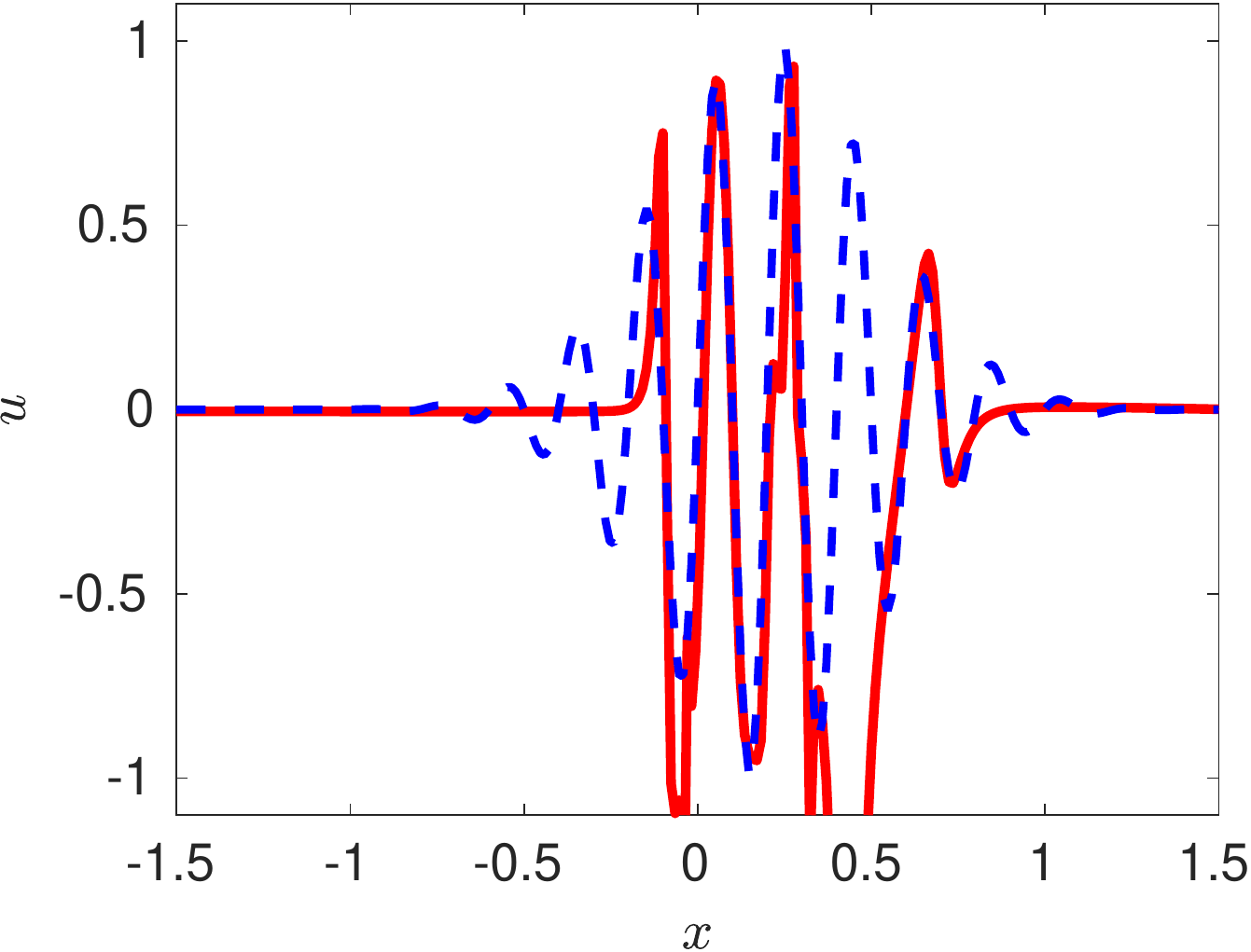}}}
			\caption{Solution of the advection equation \ref{eq:advection} obtained using PINN at three time instants: (a) t=0, (b) t=0.1 and (c) t=0.2. (solid line represents PINN solution and dashed lines represents the exact solution)}
			\label{fig:wavepacket_pinn}
		\end{center}
	\end{figure*}

	In this section, we first present the solution of the advection equation obtained using the PINN.
	In one-dimension, the advection equation has the following mathematical form:
	\begin{equation}
	\frac{\partial u(x,t)}{\partial t} + \frac{\partial u(x,t)}{\partial x} = 0,
	\label{eq:advection}
	\end{equation}
	where $u(x,t)$ is the velocity field at time $t$, along the $x$-direction.
	There are two basic motivations for choosing advection equation as a primary testbed for evaluating the PINN $-$(i) to test the ability of PINN to accurately represent complicated initial conditions and (ii) to test its ability to robustly translate a complex initial profile. Both of these characteristics are essential to retrieve a space and time continuous solution to an unsteady partial differential equation. We choose a complex initial velocity profile $u(x,t=0)$:
	\begin{equation}
	u(x,t=0) = e^{-x^2} sin(10\pi).
	\label{eq:advectionIC}
	\end{equation}
	We use the open-source PINN solver available on \url{https://github.com/maziarraissi/PINNs}. Training of the PINN is performed on 25,600 collocation points. In figure \ref{fig:wavepacket_pinn}, we show the solution obtained using PINN at three time-instants. It can be observed that the conventional PINN approach fail to even represent the initial profile at $t=0$ (equation \ref{eq:advectionIC}).  
	In the conventional PINN, the position of the collocation points is random in the absence of any residual adaptive refinement. For the complicated representation of this type, we need a larger amount of training data in the regions of sharp gradients. However, this information is not known to us apriori. The cost function of the conventional PINN acts as a regularizer but only in a global sense. Therefore it is very hard for the algorithm to capture these sharp local variations in a wide computational domain. The use of an even deeper network and a larger number of collocation points would be required to train for such complicated profiles. However, using a deeper network might restrain the training speed due to a well-known issue of vanishing gradients. 
	
\begin{figure*}[h]
		\begin{center}
			\subfigure[]{
				\resizebox*{5.3cm}{!}{\includegraphics{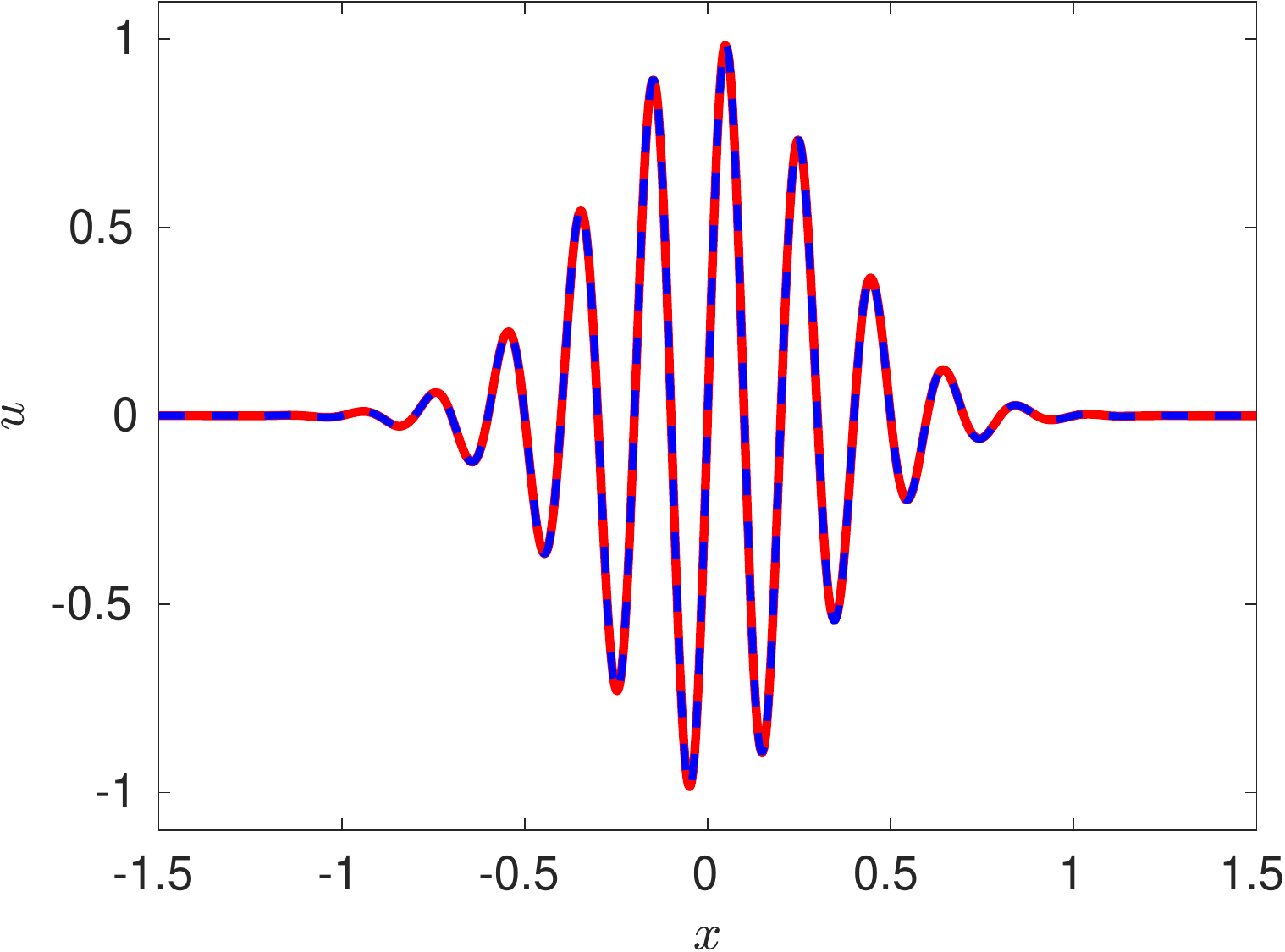}}}
			\subfigure[]{
				\resizebox*{5.3cm}{!}{\includegraphics{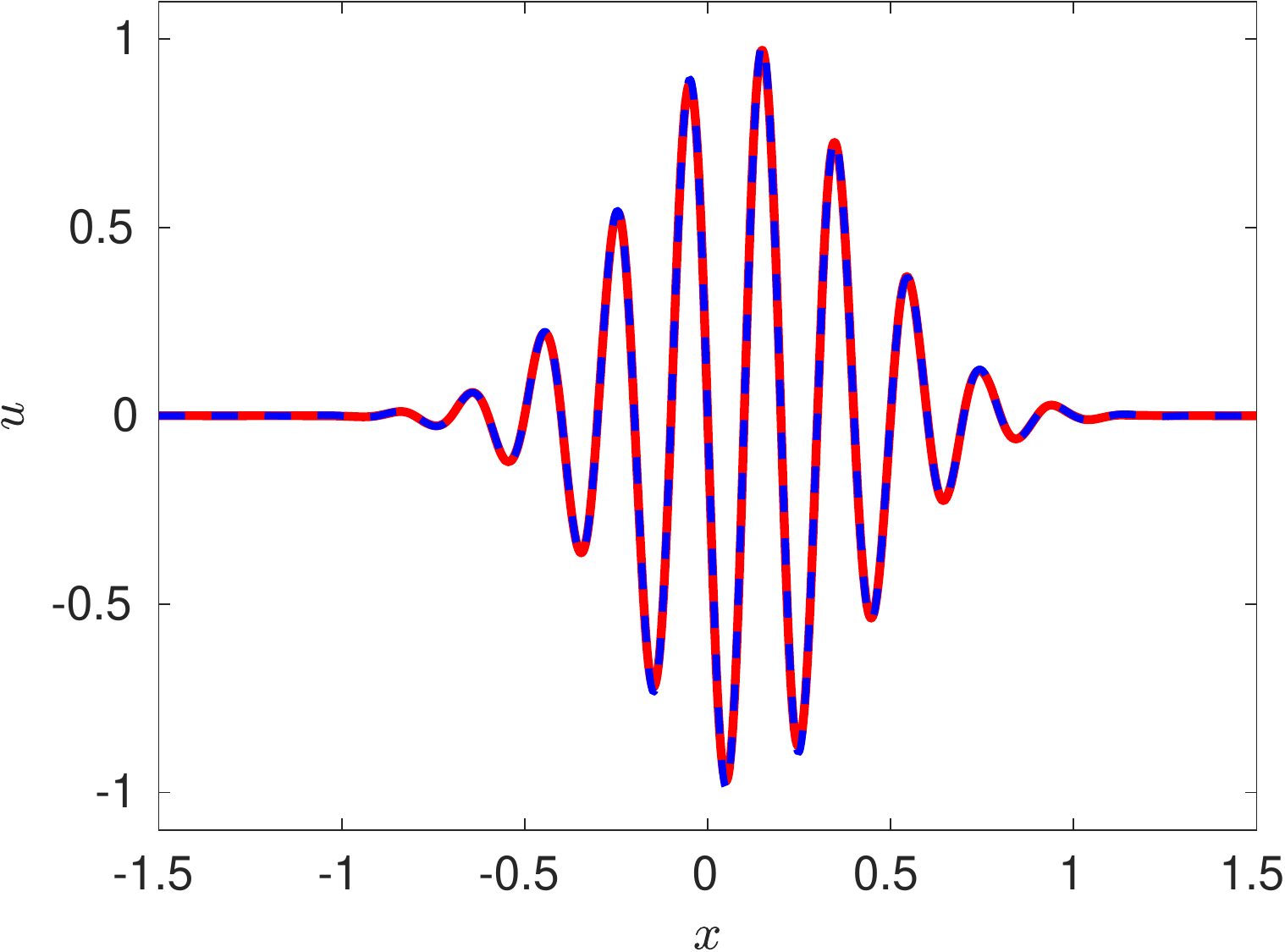}}}
			\subfigure[]{
				\resizebox*{5.3cm}{!}{\includegraphics{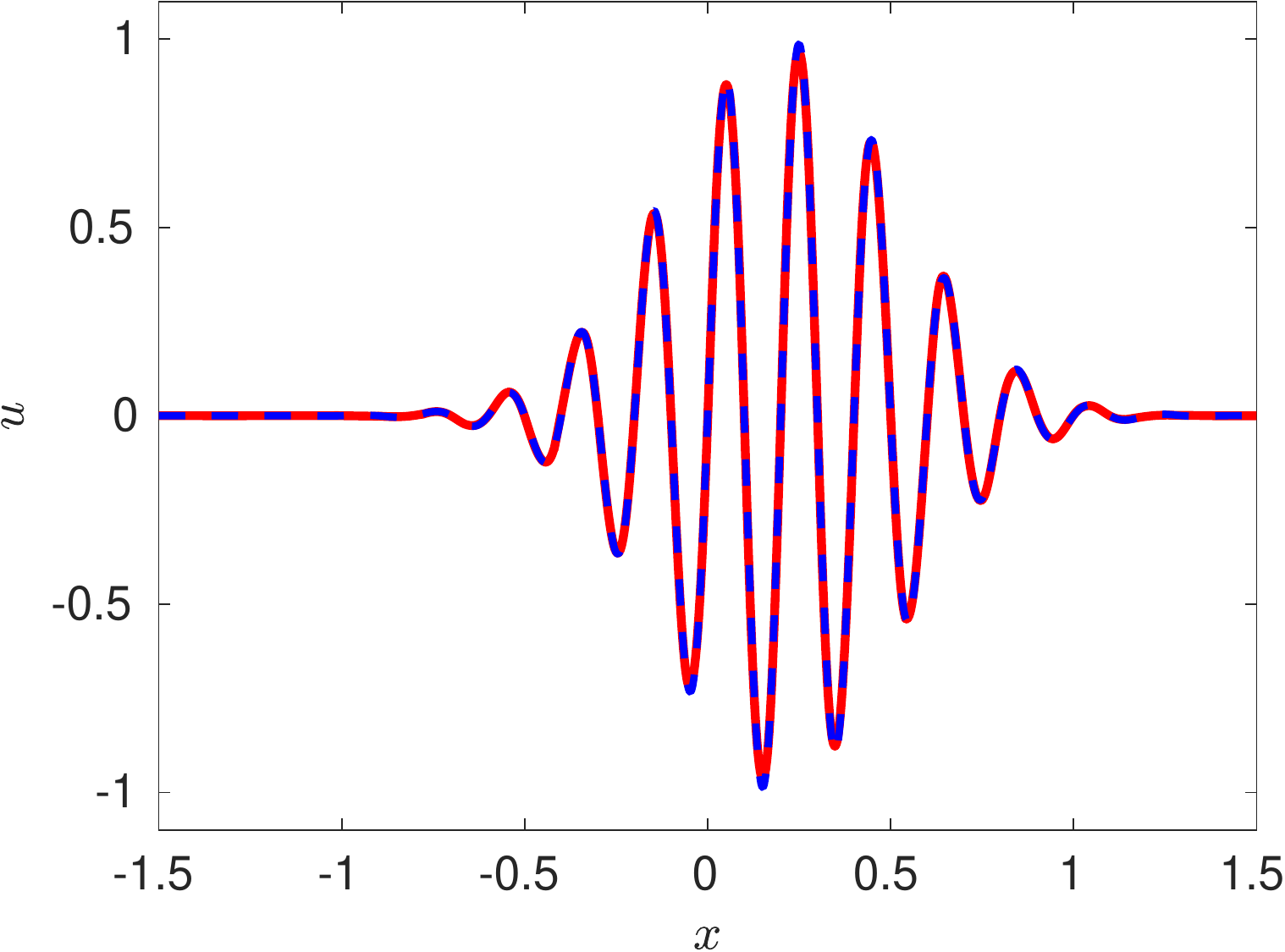}}}
			\caption{Solution of the advection equation \ref{eq:advection} obtained using DPINN at three time instants: (a) t=0, (b) t=0.1 and (c) t=0.2. (solid line represents DPINN solution and dashed lines represents exact solution)}
			\label{fig:wavepacket}
		\end{center}
	\end{figure*}

	To overcome these bottlenecks of PINN, we propose a new DPINN architecture as discussed in section \ref{s:dpinn}. The DPINN approach mitigates the above-mentioned limitations of PINN by introducing additional interface conditions, which act as a local regularization agent. This interface loss compensates for the shortage of training data in each cell. Further, the sampling of collocation points is more efficient in the DPINN architecture as compared to the original PINN as each cell now contains the same number of data points. With these motivations, we train the DPNN to solve the advection equation. The computational domain is distributed into 25 equally spaced cells along $x$-direction and five cells along $t$-axis. A neural network is employed on each of these cells. Each of these distributed neural networks is trained on 45 equally spaced collocation points using a two-layer network with five neurons in each layer. In figure \ref{fig:wavepacket} we present the solution of the advection equation \ref{eq:advection} obtained using DPINN. These plots (figure \ref{fig:wavepacket}) are obtained at three different time instants using a separate testing data-set of 4,221 unique collocation points. The mean-squared errors for the solutions presented in figure \ref{fig:wavepacket} is $\sim1.0e-05$. 
	
	In summary, the observed robustness of the proposed DPINN architecture can be attributed to the use of simpler neural networks over various segments of the computational domain. The DPINN basically tries to approximate a complicated spatio-temporal profile, by using piece-wise simpler neural networks. Such simpler neural networks are easier to train as compared to the original PINN architecture proposed by \cite{raissi2019physics}. 
	
	\subsection{Burgers' equation}
	\label{ss:burger}
	In this section, we demonstrate a direct comparison of the proposed DPINN against the original PINN in terms of the accuracy of the predicted solution. For this, we compare the solution to the Burgers' equation obtained using DPINN against that of \cite{raissi2019physics}. Burgers' equation is a non-linear partial differential equation. It is considered to be a canonical test case to demonstrate the ability of any numerical scheme to capture shocks/discontinuities in the flow field.
	The Burgers' equation in one-dimension can be expressed as:
	\begin{equation}
	\frac{\partial u(x,t)}{\partial t} + u\frac{\partial u(x,t)}{\partial x} = 0.
	\label{eq:burger}
	\end{equation} 
	\cite{raissi2019physics} employed the PINN network to find solution to the Burgers' equation corresponding to a sinusodial initial condition:
	\begin{equation}
	u(x, t=0) = sin(-\pi x).
	\end{equation}
	
	\begin{figure*}[h]
		\begin{center}
			\subfigure[]{
				\resizebox*{5cm}{!}{\includegraphics{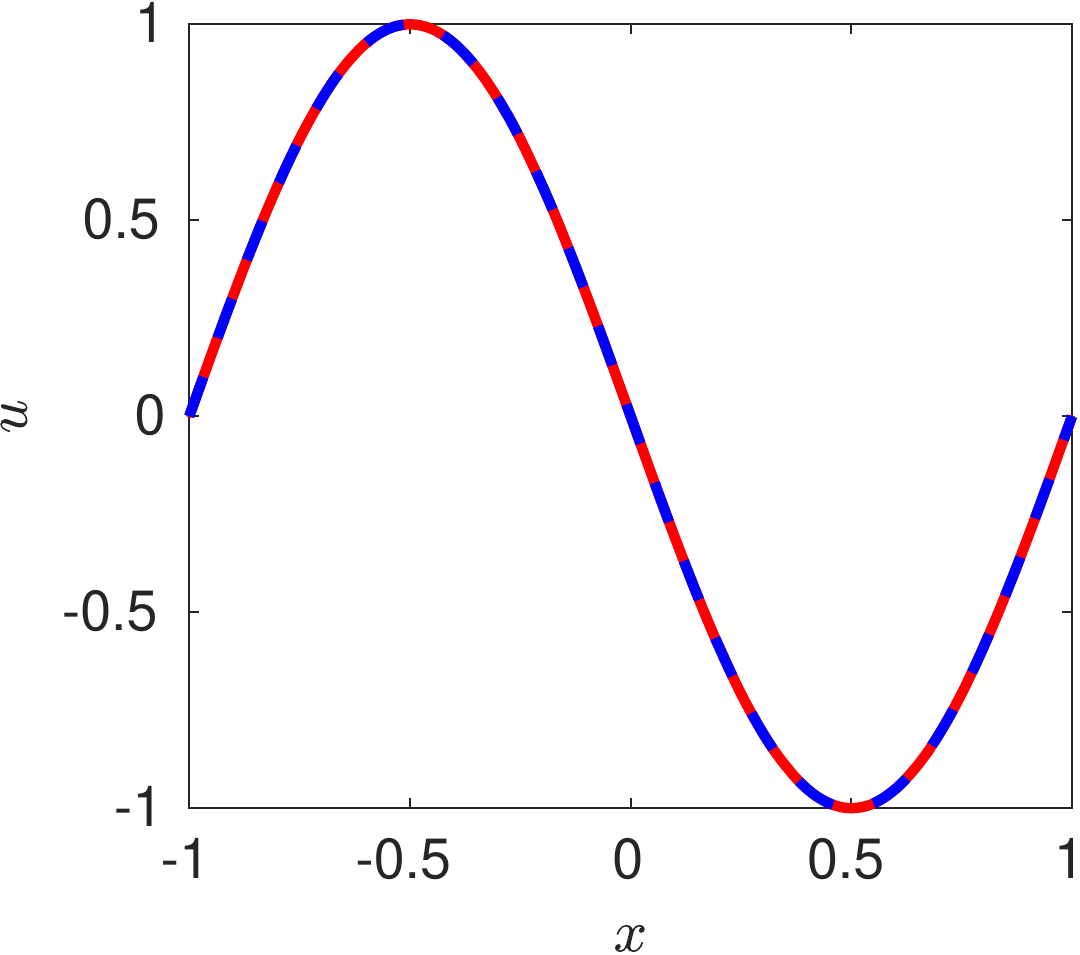}}}
			\subfigure[]{
				\resizebox*{5cm}{!}{\includegraphics{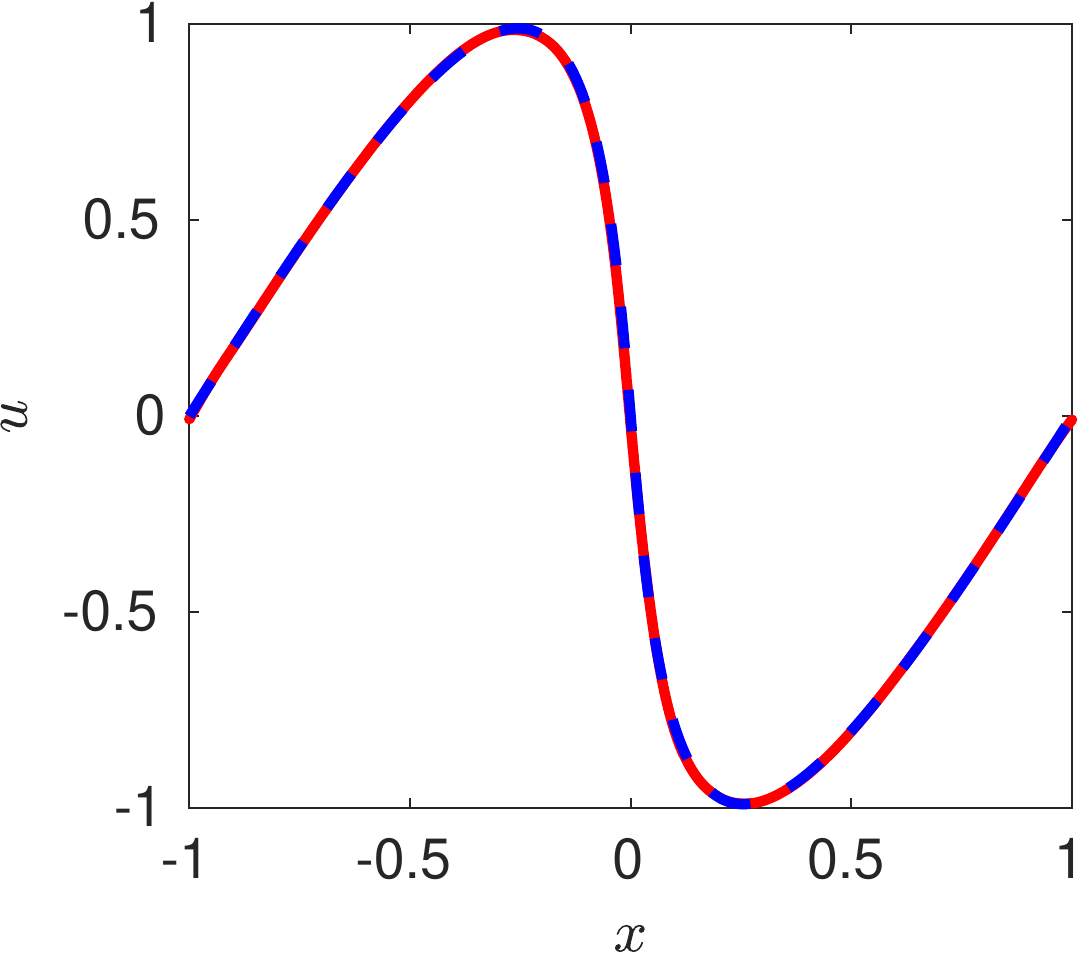}}}
			\subfigure[]{
				\resizebox*{5cm}{!}{\includegraphics{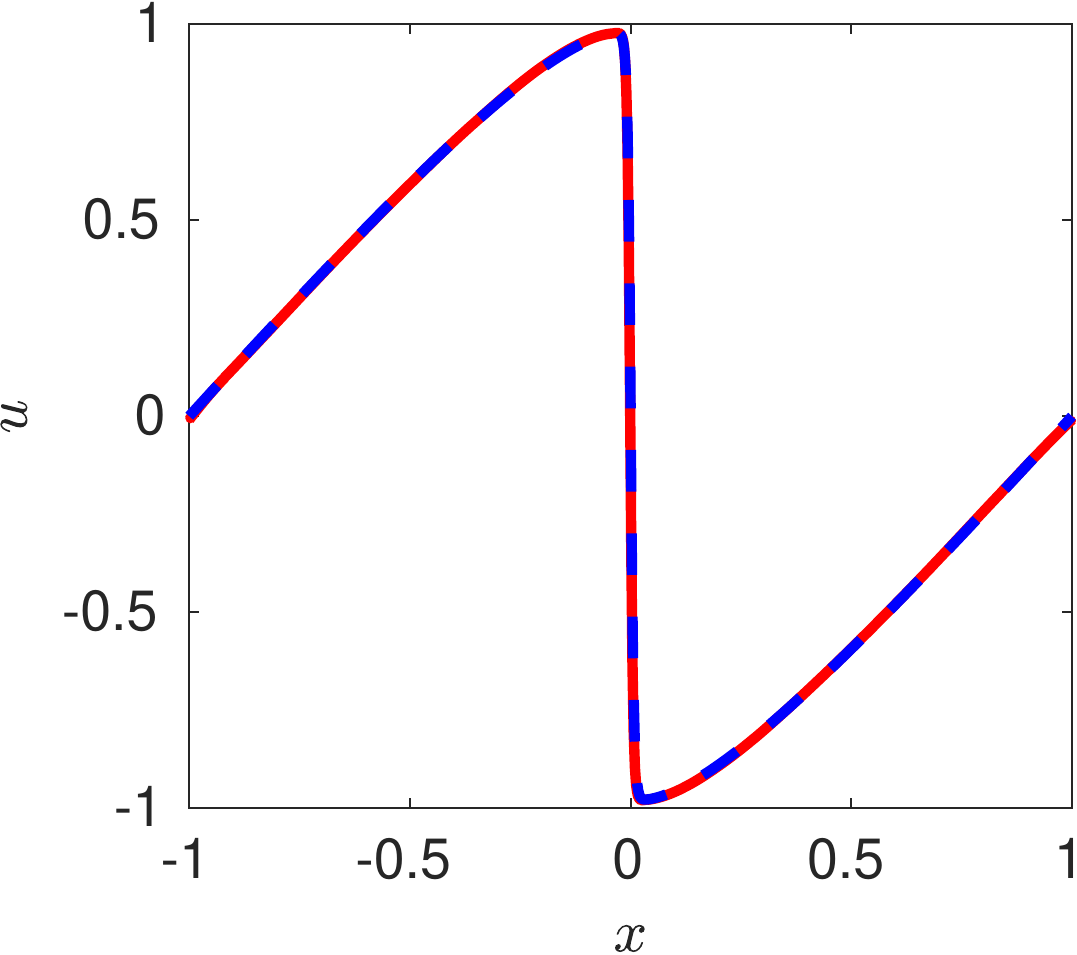}}}
			\caption{Solution of the Burger' equation \ref{eq:burger} obtained using DPINN at three time instants: (a) t=0, (b) t=0.25 and (c) t=0.5. (solid line represents PINN solution and dashed lines represents exact solution)}
			\label{fig:burger}
		\end{center}
	\end{figure*}
	
	In figure \ref{fig:burger}, we present the solution obtained at three different time instants using DPINN. The training of DPINN is performed using 4,141 collocation points. These points were distributed in 25 equally spaced cells along $x$-direction and ten equally spaced cells along $t$-axis. We use a two-layer neural network for each distributed domain with five neurons in each layer. The predicted solution (figure \ref{fig:burger}) is obtained on 1,001 collocation points with a relative L2-norm error of $2.6e$-$04$. 
	
	\cite{raissi2019physics} trained their PINN on various configurations and reported a maximum accuracy while training on $\sim 10,000$ collocation points using an eight-layer deep network with 40 neurons per layer. They reported a minimum relative L2-norm error of $4.9e$-$04$. On the other hand, by using distributed simpler neural networks DPINN achieves a smaller prediction error while training on a smaller dataset ($\sim1/2$ the data required for PINN). Hence besides being computationally more accurate, the DPINN architecture has been found to be more data-efficient as compared to the PINN architecture. 
	
	\subsection{Navier-Stokes equation}
	\label{ss:NS}
\begin{figure*}
		\begin{center}
			\resizebox*{6cm}{!}{\includegraphics{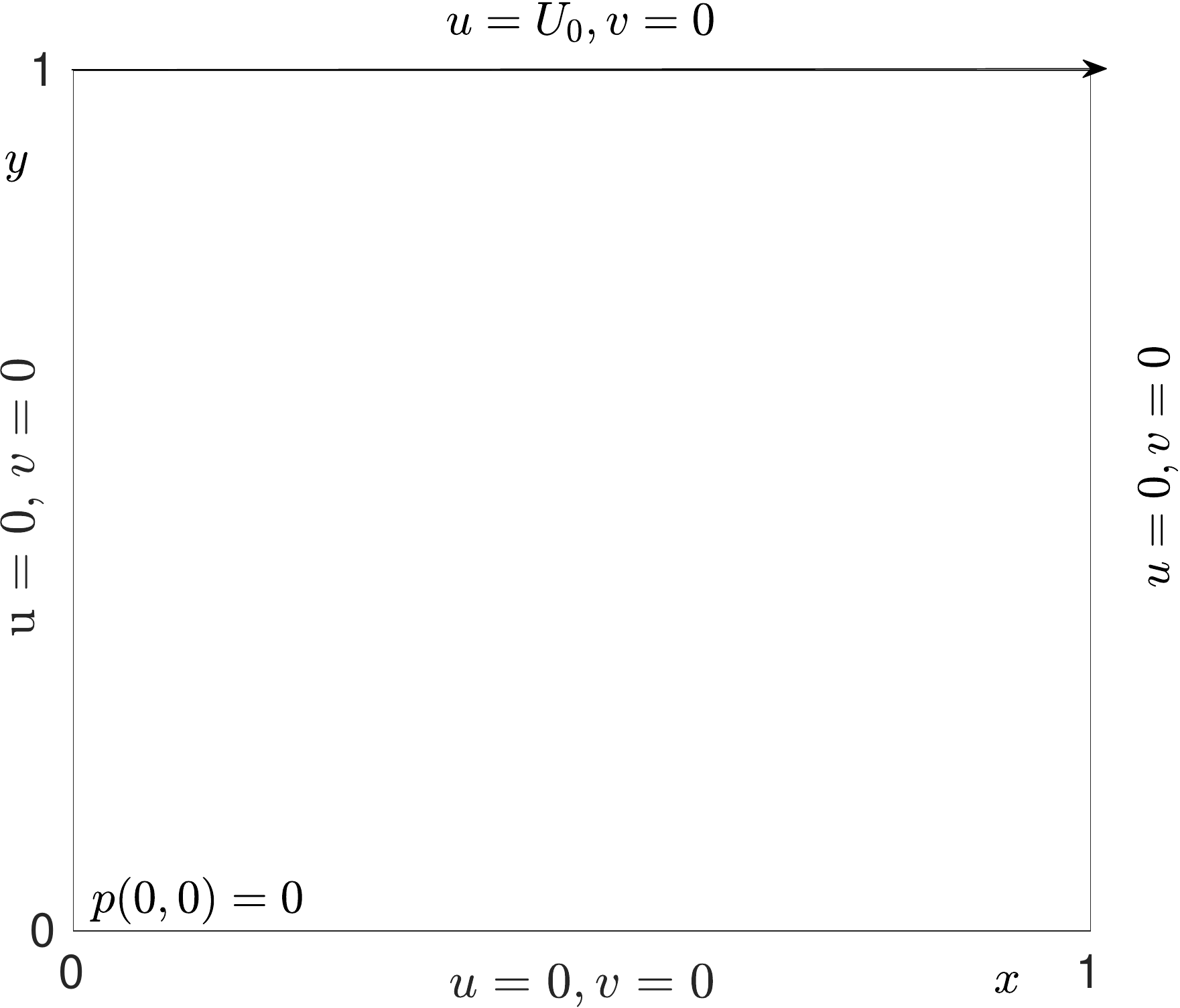}}
			\caption{Lid driven cavity problem with specified boundary conditions.}
			\label{fig:cavity}
		\end{center}
	\end{figure*}

	In this section we go a step ahead in testing the DPINN to find solution to a system of non-linear partial differential equations. The Navier Stokes equation is a system of partial differential equations and is considered to be one of the most challenging equation to solve. We choose to solve the equations to find the velocity and pressure field inside a two-dimensional square cavity with a moving top plate (lid driven cavity problem), as shown in figure \ref{fig:cavity}. At a smaller Reynolds number, this flow is known to possess a incompressible steady state solution. Hence, we can directly simplify the Navier stokes equation, to yield the steady state solution by removing the temporal term. Further, we impose the divergence-free incompressibility condition to arrive at the following form:
	\begin{eqnarray}
	\frac{\partial{u}}{\partial{x}}+\frac{\partial{v}}{\partial{y}}&=&0;
	\label{eq:con}\\
	u\frac{\partial{u}}{\partial{x}} + v\frac{\partial{u}}{\partial{y}}&=&-\frac{1}{\rho}\frac{\partial{p}}{\partial{x}}+\nu\frac{\partial^2 u}{\partial{x^2}} + +\nu\frac{\partial^2 u}{\partial{y^2}};
	\label{eq:mom_x}\\
	u\frac{\partial{v}}{\partial{x}} + v\frac{\partial{v}}{\partial{y}}&=&-\frac{1}{\rho}\frac{\partial{p}}{\partial{y}}+\nu\frac{\partial^2 v}{\partial{x^2}} + +\nu\frac{\partial^2 v}{\partial{y^2}};
	\label{eq:mom_y}
	\end{eqnarray}
	where, u and v are the velocity component along the x and y direction, $p$ is the pressure, $\nu$ represents the kinematic viscosity and $\rho$ is the fluid density. Equation (\ref{eq:con}) is the continuity equation, while equations (\ref{eq:mom_x}) and (\ref{eq:mom_y}) represents the momentum equations.
	\begin{figure*}[ht!]
		\begin{center}
			\subfigure[]{
				\resizebox*{5.3cm}{!}{\includegraphics{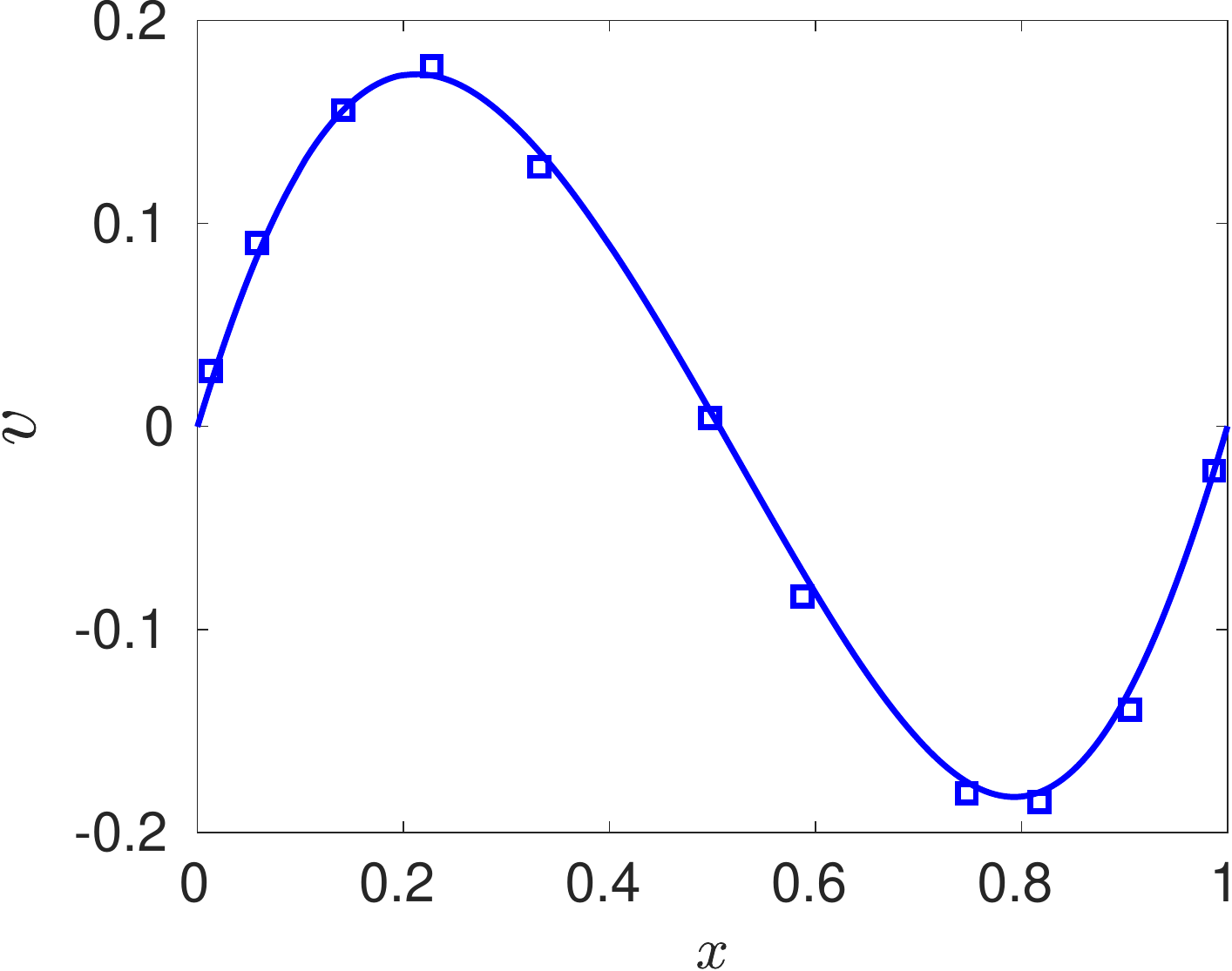}}}
			\subfigure[]{
				\resizebox*{5.3cm}{!}{\includegraphics{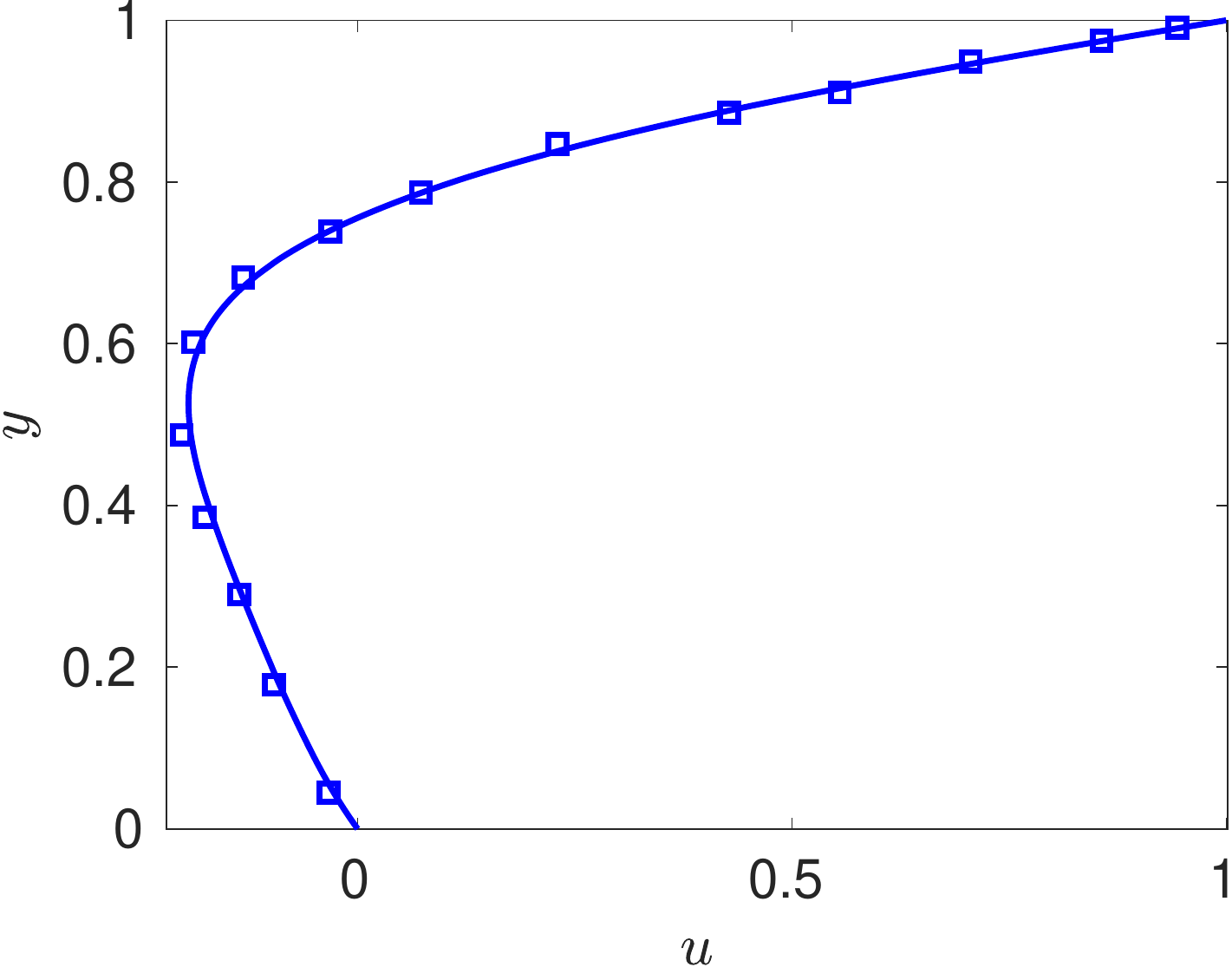}}}
			\subfigure[]{
				\resizebox*{5.2cm}{!}{\includegraphics{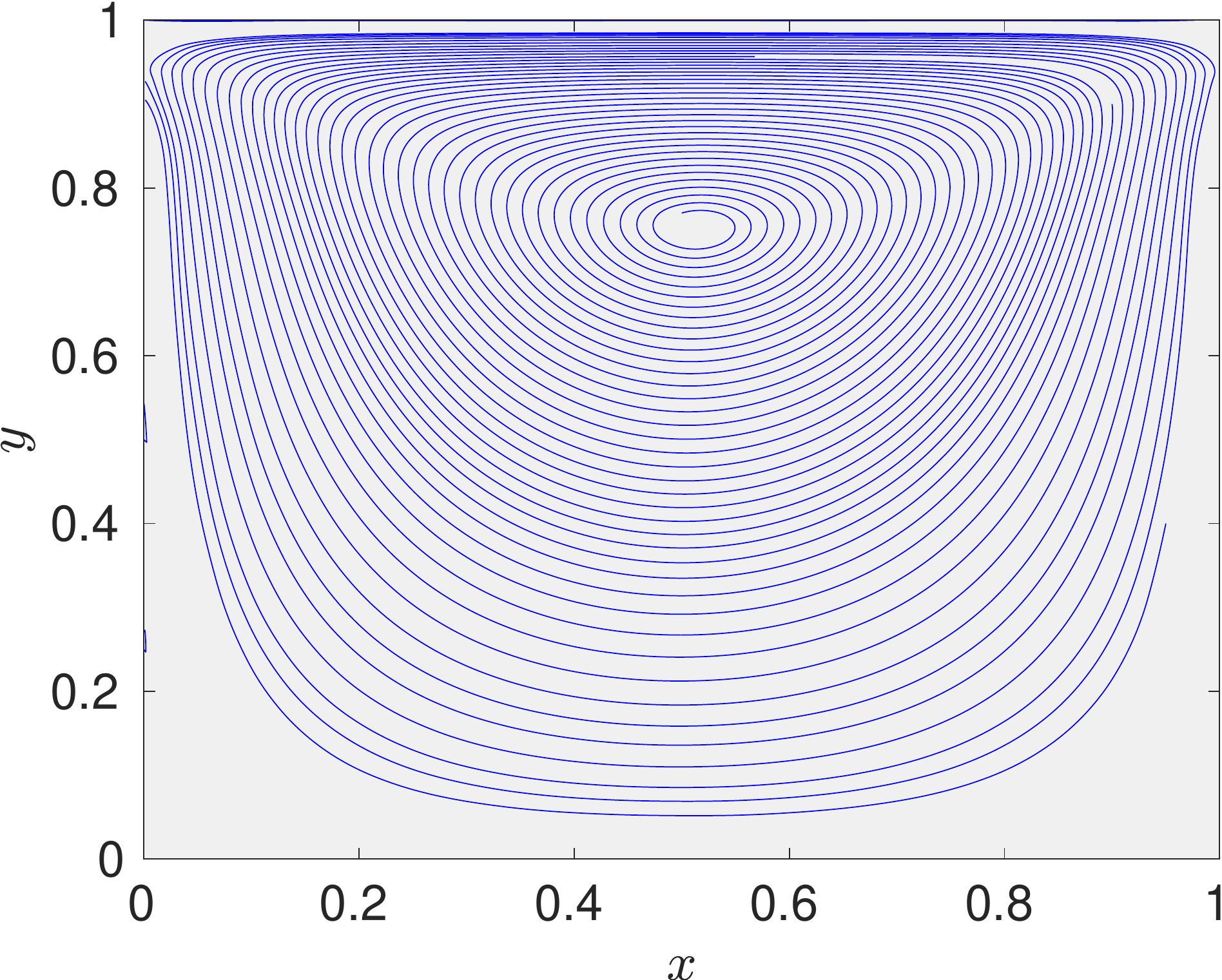}}}
			\subfigure[]{
				\resizebox*{5.35cm}{!}{\includegraphics{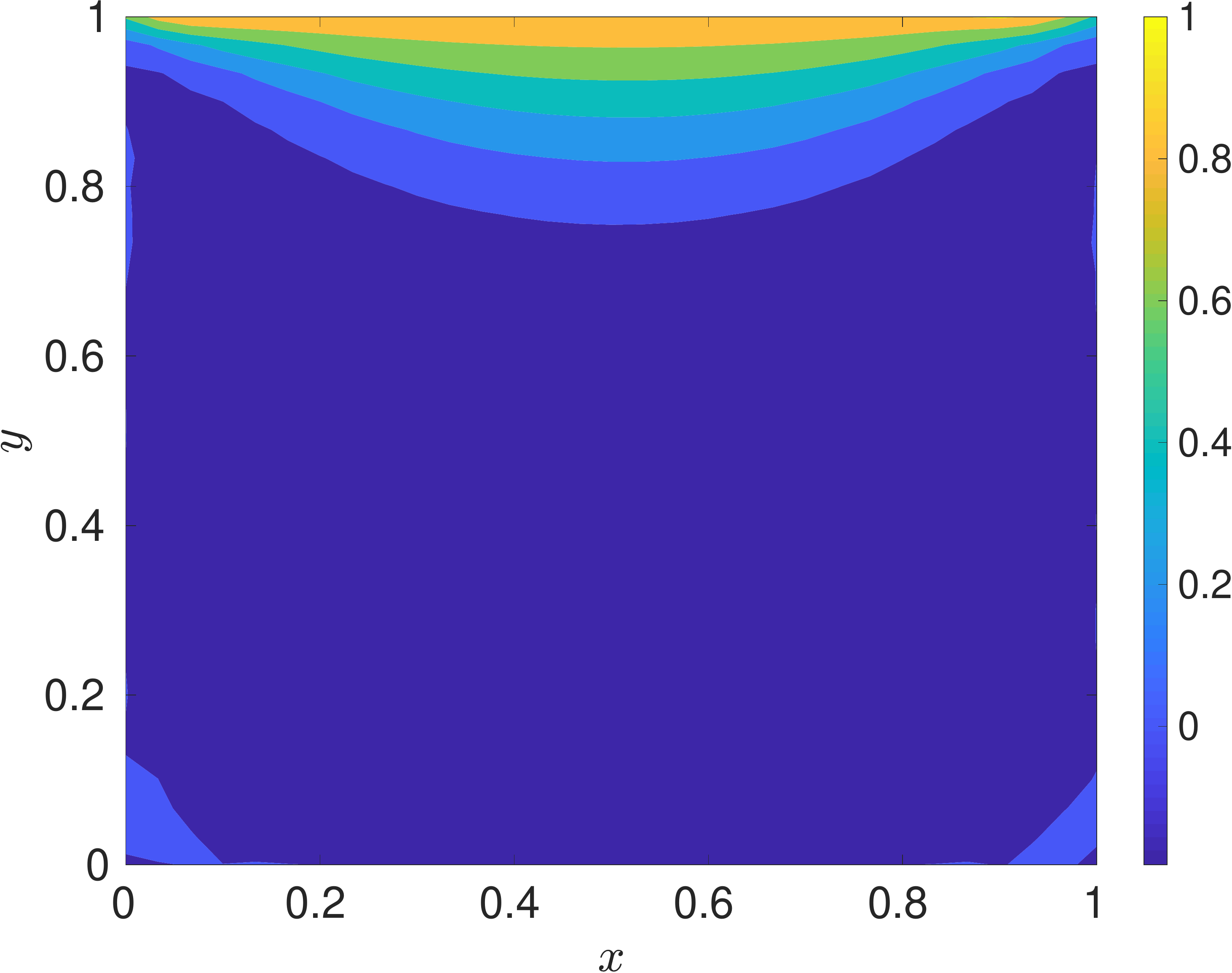}}}
			\subfigure[]{
				\resizebox*{5.35cm}{!}{\includegraphics{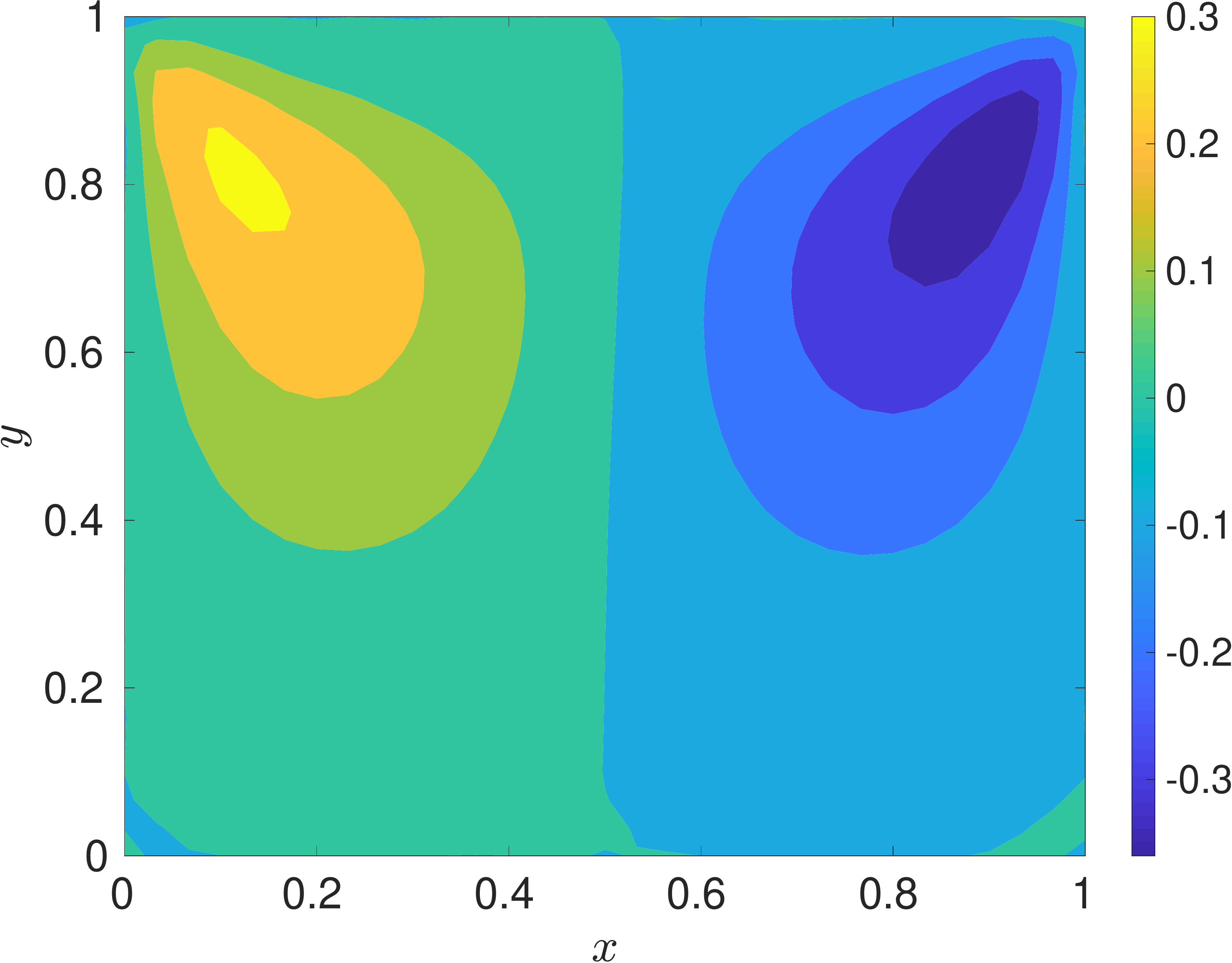}}}
			\subfigure[]{
				\resizebox*{5.25cm}{!}{\includegraphics{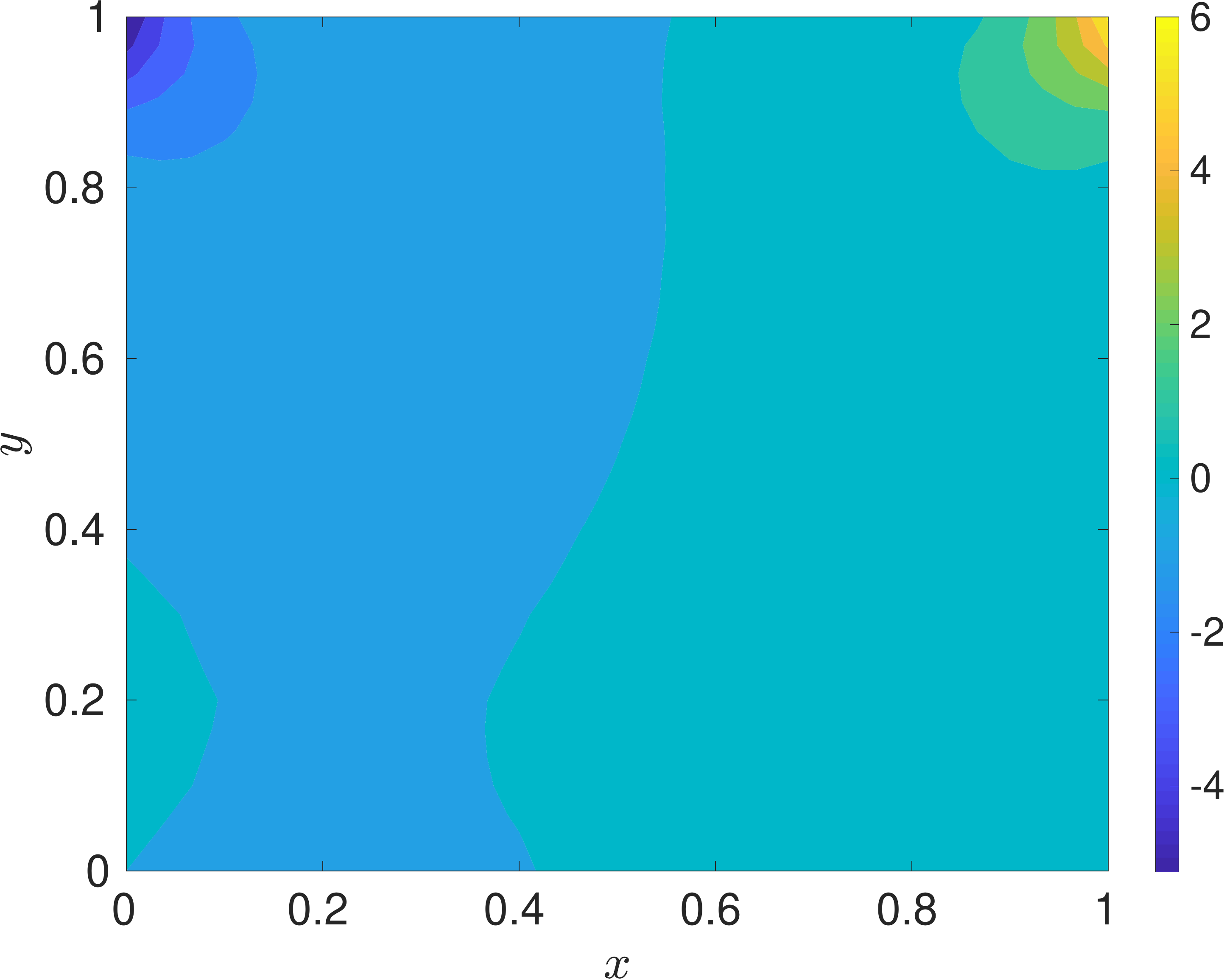}}}
			\caption{Lid-driven cavity steady state solution predicted by the DPINN: (a) $v$-velocity at centerline $x=0.5$, (b) $u$-velocity at centerline $y=0$, (c) streamlines, (d) $u$-velocity contours, (e) $v$-velocity contours and (f) pressure contours.}
			\label{fig:lid_cavity}
		\end{center}
	\end{figure*}
	Although these equations have been simplified, they still are, in their presented form, a highly non-linear system of parital differential equations. To the best of our knowledge, this is the first such attempt to directly solve the Navier-Stokes equation directly using a PINN based approach.
	
	Since, for the problem of interest, the flow is incompressible, density can be assumed to be constant. Hence, we would train our proposed DPINN to retrieve the pressure and velocity field only. The problem domain and the boundary conditions are specified in figure \ref{fig:cavity}. The Reynolds number of the flow is 10 ($Re = \rho U_0 L / \nu = 10$, where L is length of the plate).
	The training of the DPINN is performed using 961 collocation points. These points are distributed in ten equally spaced cells along $x$-direction and $y$-direction. We use a two-layer neural network for each distributed domain with five neurons in each layer. In figure \ref{fig:lid_cavity}, we show the predictions obtained using DPINN. The $v$-velocity obtained at the centerline $x=0.5$ is presented in \ref{fig:burger}(a) and $u$-velocity obtained at the centerline $y=0.5$ is presented in \ref{fig:burger}(b). Both of these results are compared against well-established results of \cite{ghia1982high}. It can be observed that both the centerline velocities obtained using DPINN are in good agreement with the results of \cite{ghia1982high}. Further, in figure \ref{fig:lid_cavity}(c) we show the streamlines and in figures \ref{fig:lid_cavity}(d), \ref{fig:lid_cavity}(e) and \ref{fig:lid_cavity}(f) we show velocity and pressure contours obtained using DPINN. These contours and streamlines depict the physical flow pattern that is expected to form for this flow problem.
	
	Although, for the presented problem, DPINNs have been able to find the solution to the Navier-Stokes equation. We want to clarify that, the flow problem simulated in this work (lid-driven cavity) has a low Reynolds number. With increasing Reynolds number, the complexity of the solution increases since the flow becomes unsteady, turbulent and three-dimensional. The current work sets the foundation for the use of DPINNs to directly solve the Navier-Stokes equations and is, in fact, the first such attempt in this direction.
	
	\section{Conclusions}
	\label{s:summary}
	In this paper, we present DPINN$-$a data-efficient distributed version of PINN to solve linear as well as nonlinear PDEs. The proposed DPINN enhances the capability of PINN by improving both its network architecture and the employed learning algorithm. DPINN incorporates a divide-and-conquer type strategy analogous to finite volume methods by partitioning the computational domain into smaller sub-domains (called cells) and installing a local PINN in each of these cells. This partitioning breaks the hard problem which potentially requires a very deep PINN, into smaller sub-problems which can be solved by various minimal sized local PINNs. The additional physical constraints at the interfaces act as natural network regularizers, which boosts the representation capability of the network. The cost function of DPINN itself stitches the individual local PINN solutions, which makes DPINN even more data-efficient than the original PINN.

	The major highlights of this study are as follows:
	\begin{enumerate}
		\item We have proposed a novel improved PINN algorithm (called DPINN) which is more data-efficient and addresses the vanishing gradient issue encountered by deep PINNs.
		\item We provide a glimpse of the improved representation power of DPINN by employing it to solve the advection equation. We find that while PINN fails even to represent a high-frequency wave packet, DPINNs not just accurately represents the wave packet, but robustly advects it as well.
		\item We show that DPINN yields a more accurate solution to the Burgers' equation with a smaller amount of training data as compared to the PINN solution.
		\item  We show that DPINNs can be used to solve steady-state Navier-Stokes equations at low Reynolds number. However, to solve unsteady full three-dimensional Navier-Stokes equations, a more detailed analysis needs to be performed. However, such a study will require more advancements in the algorithm and larger computational resources, which is beyond the scope of the present work.  
	\end{enumerate}
	
	This paper has demonstrated that DPINNs can be efficiently used to find data-driven solution to PDEs. The next obvious task in this direction is to extend the DPINN approach for the inverse problem, ie. data-driven discovery of PDEs. We are currently working in this direction and will report consequent progress in our future work.

	\section*{References}
	\bibliographystyle{elsarticle-harv}
	\bibliography{mybibfile}
	
\end{document}